# A Machine Learning-Driven Solution for Denoising Inertial Confinement Fusion Images

Asya Y. Akkus, Bradley T. Wolfe, Pinghan Chu, Chengkun Huang, Chris S. Campbell, Mariana Alvarado Alvarez, Petr Volegov, David Fittinghoff, Robert Reinovsky, Zhehui Wang

***Los Alamos National Laboratory; Bikini Atoll Rd, Los Alamos, NM 87544***

***Lawrence Livermore National Laboratory; 7000 East Ave, Livermore, CA 94550***

*Abstract*—**Neutron imaging is essential for diagnosing and optimizing inertial confinement fusion implosions at the National Ignition Facility (NIF). The Neutron Imaging System (NIS) requires 10-µm resolution to feasibly perform reconstruction through iterative algorithms [1]. Additionally, low-yield source images are often degraded by various types of noise. Gaussian and Poisson noise often coexist within one image, obscuring fine details and blurring the edges where the source information is encoded. Traditional denoising techniques such as filtering and thresholding can inadvertently alter critical features or reshape the noise statistics. This significantly impacts the ultimate fidelity of the downstream iterative image reconstruction pipeline. However, recent advances in synthetic data production and machine learning have opened new opportunities to address these challenges. In this study, we present an unsupervised autoencoder with a Cohen-Daubechies-Feauveau (CDF 97) wavelet transform in the latent space designed to suppress for mixed Gaussian-Poisson noise while preserving essential image features. The network successfully denoises neutron imaging data. Benchmarking against both simulated and experimental NIF datasets demonstrates that our approach achieves lower reconstruction error and superior edge preservation compared to conventional filtering methods such as Block-matching and 3D filtering (BM3D). By validating the effectiveness of unsupervised learning for denoising neutron images, this study establishes a critical first step towards AI-driven, end-to-end reconstruction frameworks for ICF diagnostics.**

## I. INTRODUCTION

INERTIAL confinement fusion (ICF) is the process of compressing fusion fuel at high temperatures and pressures. Subsequently, the fuel's inertia confines it for the brief duration necessary for two nuclei to overcome Coulomb repulsion and fuse [2,3]. Most commonly, ICF uses deuterium-deuterium (DD) or deuterium-tritium (DT) fusion reactions, producing energetic neutrons. ICF targets can be driven by direct or indirect drive depending on the technical objectives of a given experiment [4,5]. Direct drive ICF relies on multiple lasers directly heating, ablating, and compressing a small capsule containing fusion fuel. In indirect-drive ICF, lasers heat a hohlraum containing the capsule, and radiation from the hohlraum then heats, ablates and compresses the capsule [2,6].

Neutron imaging is a fundamental diagnostic used in ICF experimental campaigns. The variation in experimental conditions across diverse yields and corresponding signal-to-noise conditions necessitates robust analysis methods [2]. Neutron imaging systems employ pinhole and penumbral apertures to measure the spatial morphology of the fusion burn region [6]. Pinhole apertures are smaller in size than the source, while penumbral apertures are larger than the source. Information captured during the ICF event is primarily found within the penumbral shadow of the aperture, which is more diffuse and indirect. In contrast, pinhole apertures provide a clearer, more direct image of the source. Fusion-produced neutrons are easily able to penetrate dense materials. As a result, thick apertures are required at NIF to create sufficient image contrast. However, adequate contrast remains highly condition dependent. For shots with low yield or small hot spot sizes, image processing and reconstruction techniques are required to analyze data and perform 3D reconstructions [6]. In this study, the NIF images used for denoising analysis were captured from plates that collected neutrons passing through thick pinhole and penumbral apertures. As the aperture projections are not direct representations of the source, imaging the thermonuclear hotspot requires complex reconstruction techniques that are vulnerable to noise.

While pinhole imaging has many advantages when compared to neutron imaging from other apertures, certain challenges to visual data quality remain [7]. The low solid angle of pinhole apertures limits their ability to image certain smaller ICF cases appropriately due to decreased signal strength and neutron concentration per pinhole [6]. Additionally, pinhole images are more susceptible to noise when compared to annular or ring apertures [6]. Particularly in high-yield neutron shots, there are

**This work was conducted under the auspices of the DOE SULI and Los Alamos National Laboratory LDRD programs under project 20250870ER.**

numerous sources of noise [5]. Poisson shot noise is produced due to discrete photons or particles colliding with the detector and is commonly present in low-light and low-yield conditions. Poisson noise becomes Gaussian when the photon and particle count generated by the ICF event is higher than average due to the presence of electronic sensor noise. Uniform noise, much like Poisson noise, is seen in situations where lighting is poor. Salt-and-pepper noise arises due to transmission errors or sensor malfunctions. Finally, defects or scratches present on the plate or detector can directly affect the image, causing visible artifacts in those areas.

There are a wide variety of yield-dependent noise combinations possible in ICF neutron imaging. However, in the case of lower-yield shots used in this study, independent image analysis confirmed previous findings that Poisson noise, Gaussian noise, and plate defects are the most common sources of noise to be addressed. After different particle conversions, gain stages, and background subtractions, the noise is closer to Gaussian noise than Poisson or combined Gaussian-Poisson noise in certain NIF images. However, in ICF pinhole shots with low neutron yield, Poisson shot and Gaussian noises often combine [6]. This necessitates a need for extensive work in analyzing and mitigating the noise present in pinhole images in a manner that retains the basic properties of the original image. Most specifically, the focus of this study lies in tackling edge preservation and source reconstruction for NIF data outputs.

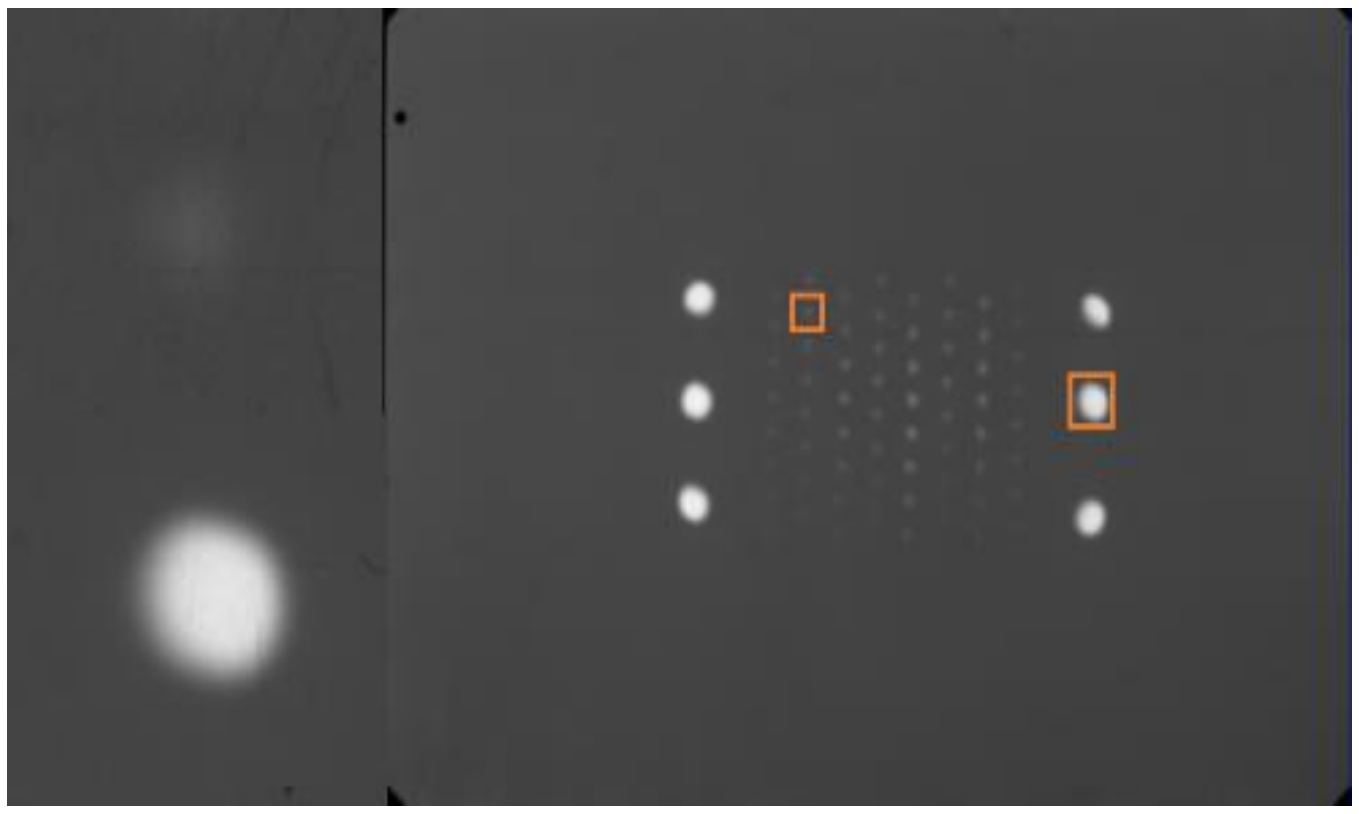

**Fig.1.** Neutron image data emitted during the N210808 shot. Neutrons produced during fusion are uncharged and highly penetrative, allowing them to travel to a scintillator, where they interact and generate visible light. The full plate for an example shot, N210808, is displayed on the right. On the left is an example of a pinhole image (top left) and penumbral image (bottom left). These are samples for the autoencoder model and were taken through region of interest (ROI) extraction of individual penumbra and pinholes.

Currently, neutron plate images are not denoised before 3-dimensional reconstruction. The generalized expectation maximization algorithms used work directly on the data [8]. In other industrial areas, methods such as Weiner filtering and BM3D are used [9,10]. In recent years, there has been a surge in Convolutional Neural Network (CNN) usage for denoising real and simulated neutron pinhole images [11]. While unsupervised methods have been explored more frequently in the recent past due to their small-training-set robustness, they have not been popular in the computational ICF research space due to the complex noise environment and data paucity [12].

The current lack of ground truth data in ICF neutron imaging stems from two main factors. Firstly, there is a relatively limited body of experimental data from the NIF, making it difficult to train and test large machine learning models. Additionally, there is a lack of high-resolution images that training sets can utilize to learn patterns. Because optimal supervised models must use simulated data conservatively in tandem with real-world ground truths to generate accurate results, the development and validation of unsupervised testing methods is necessary [10].

Methodologies such as clustering and unsupervised autoencoders can better leverage inherent features of the noisy images rather than relying on potentially inaccurately simulated ground truth datasets during training. Additionally, unsupervised networks require fewer simulated datasets post-tuning, and less validation compared to supervised networks. While unsupervised machine learning methodologies have their own unique challenges with reconstruction error, a combination of unsupervised and filtering methodologies has not been significantly explored as an alternative. A literature search only yielded one study on this, where a fast Fourier transform was employed as a custom layer in a CNN [13].

This publication explores the use of an unsupervised denoising autoencoder with a discrete CDF 97 wavelet transform in the latent space to denoise and sharpen neutron pinhole images. The CDF 97 biorthogonal wavelet transform has been shown to be highly effective due to its excellent energy compaction and smooth basis functions, which enable efficient separation of signal and noise components [14]. Model effectiveness was benchmarked using synthetic ground truth data created by a forward model which was then corrupted by Gaussian and Gaussian-Poisson noise. These models produced results representative of experimental NIF data under different neutron yields. The autoencoder's performance relative to this synthetically generated ground truth was then compared with gold-standard methodologies such as BM3D, Gaussian, and Weiner filtering. The model was able to outperform the latter methods for edge preservation metrics such as polar and cartesian coordinate visualization, radial position of the edge, spread of the edge at transitions, residuals, and radial profile. Tikhonov regularization was also applied to evaluate how effectively the reconstruction recovered the original source.

## II. Materials and Methods

An unsupervised autoencoder with custom layers in the latent space was utilized to denoise pinhole and penumbral images from the NIF dataset. Its performance was assessed using a forward model generating a sample of neutron images. Subsequently, the quality of the simulated ground truth reconstructions was assessed through various edge preservation metrics.

Three classes of datasets were used to validate the functionality of the autoencoder. The first two contained data simulated using a forward model following a nonlinear Gaussian and Gaussian-Poisson noise model, respectively. These datasets were used for initial benchmarking and validation. They provided quantitative assessments of the model's performance on certain noise profiles. Following successful validation on forward model data, The NIF experimental data was taken from 2 plates captured from shot N210808 as shown in Fig. 1 and the training model was applied to it.

Each plate had 64 pinholes each yielding a total available dataset of 128 images. The plate data was then split into smaller regions of interest (ROIs) of size (256, 256) pixels to be passed through the autoencoder as separate datapoints. This dataset was used to prototype the preliminary autoencoder model and served as proof of concept for the feasibility of denoising work using machine learning methods.

The benchmarking datasets used for validating autoencoder functionality were simulated through a forward model that generated a noiseless, simulated ground truth sample of a pinhole or penumbral image. As demonstrated in Fig. 2, the model consisted of two key elements convolved together to produce the simulated ground truth. The Gaussian blur simulates a source image (bottom left of Fig. 2), while the disc represents a pinhole or penumbral aperture depending on its size (top left of Fig. 2). To simulate a pinhole, the disc was made to be smaller than the source image. To simulate a penumbral image, the disc was made larger than the source image.

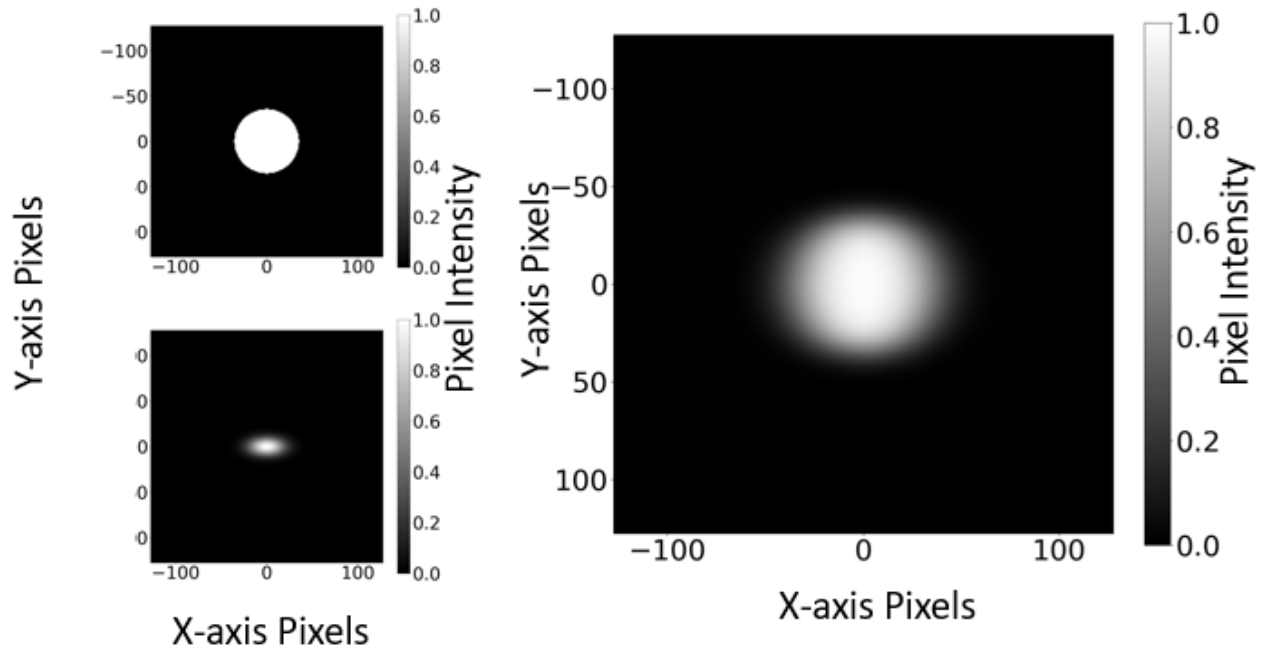

**Fig. 2.** Illustration of the forward model used for benchmarking. This setup was employed to simulate a penumbral image, which served as the central focus of the study. The simulated penumbra (top left) was convolved with the modeled source distribution (bottom left) to generate a simulated ground truth image of the penumbral aperture (right).

Two datasets representing the noise commonly seen in neutron shots were generated from the forward model. One dataset had exclusively signal-dependent Gaussian noise added to 3000 images, a sample of which is shown at the top of Fig. 3. The noisy image $I_n$ may be expressed as:

$$I_n = I_0 + (RMS(I_0) \times \varepsilon) \quad (1)$$

Here $I_0$ is the clean image, and $\varepsilon$ represents Gaussian noise with a variance of 1.

The other had a combination of Gaussian and Poisson noise calculated by equation 2, and as shown in the bottom image of Fig. 3. To mirror conditions at the NIF, the SNR is 10 and $\varepsilon$ is Gaussian noise with the same parameters as the previous model [1].

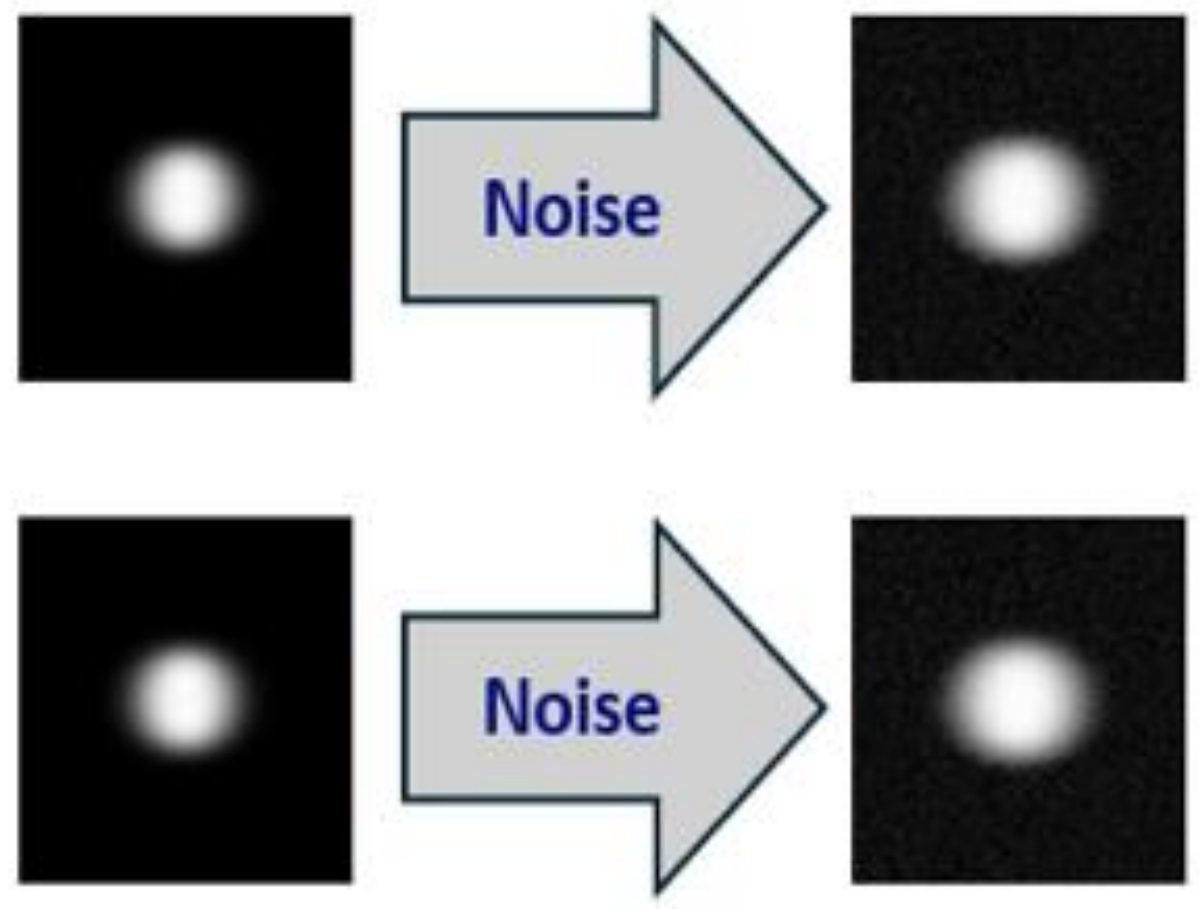

**Fig. 3.** A sample from the forward model image from Fig. 2a with Gaussian noise (top) and mixed Gaussian-Poisson noise (bottom) added.

$$I_{n+1}(x, y) = \text{Poisson}\left(\frac{SNR^2 * I_n(x, y)}{\max(I_n)}\right)\frac{\max(I_0)}{SNR} \quad (2)$$

The autoencoder was built using the PyTorch library and consists of 3 main components: an encoder, a linear transformation layer in the case of penumbral image denoising, a custom CDF 97 discrete wavelet transform layer, and a decoder. As shown in Fig. 4, the image undergoes a series of dimensional transformations as it propagates through the layers of the deep learning model.

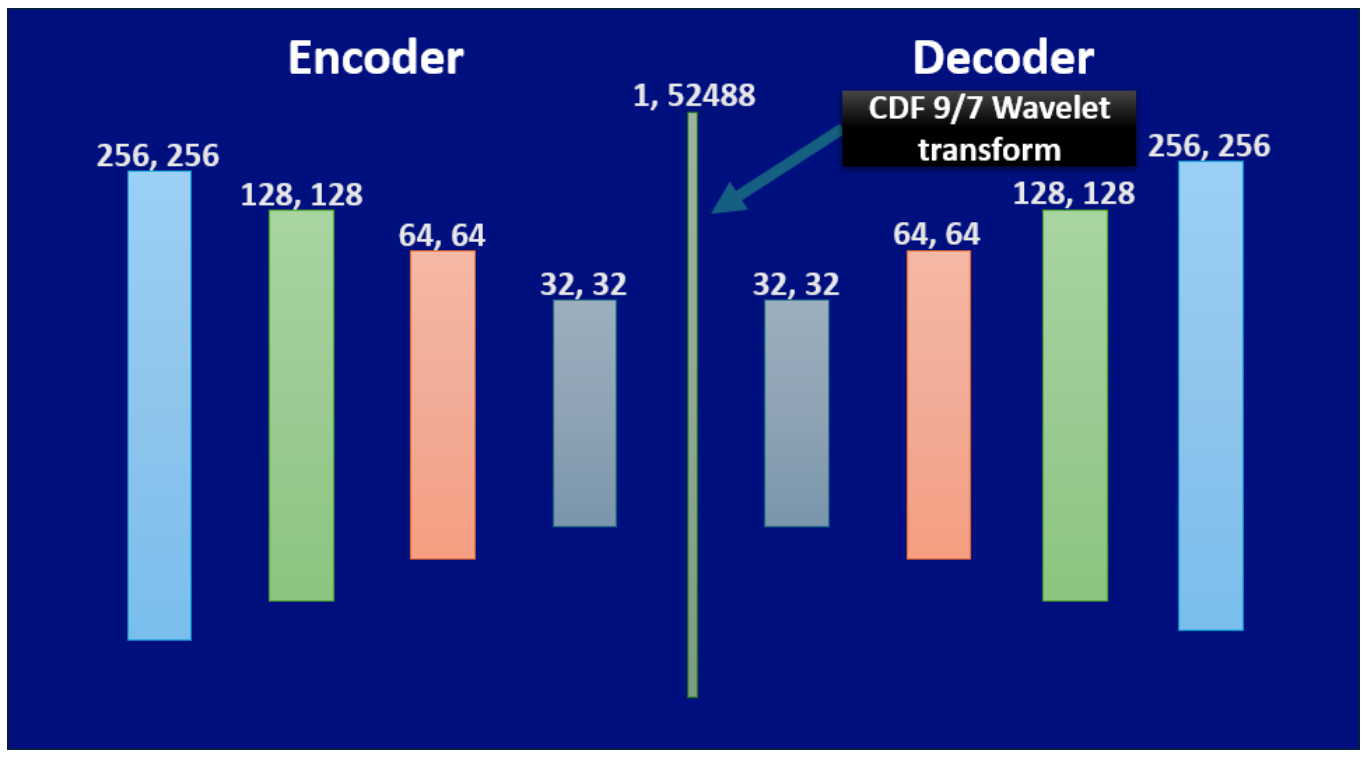

**Fig. 4.** An illustration of the dimensional transformations applied to each image as it passes through the stages of the autoencoder. The repeated minimization of dimensions within the encoder portion allows for the identification and preservation of key features and the removal of noise vectors. The decoder is then able to reconstruct the denoised image using the most dominant features, while the CDF 97 wavelet transform removes any noise that was missed during the encoding and flattening process.

The encoder extracts the most relevant features from the inputs by downsampling each image individually. This process consists of six two-dimensional convolutional layers, each accompanied by leaky ReLU activation function layers that introduce non-linearity during training. This allows the model to learn image features more effectively when compared to linear processes such as Gaussian and Weiner filtering [15]. Meanwhile, the convolutional layers reduce the spatial resolution of the input image by half each time with a kernel size of 5x5, a stride of 2, and a padding value of 2. This combination of kernel size, stride, and padding results in reduction of image dimensions while maintaining the number of feature maps at each layer to ensure that the autoencoder has sufficient information to learn from during the training process. The first and second convolutional layers were followed by pooling layers after the activation function. The third, fifth, and sixth convolutional and leaky ReLU function pairs were followed by a dropout layer. The fourth convolutional layer was followed by a dropout layer and pooling layer. Dropout layers were set to 0.2. They were utilized to help prevent overfitting by randomly zeroing twenty percent of the input units at each layer during each forward pass in the training loop. The pooling layers used had a window size of 2x2 and worked to reduce spatial dimensions even further to ensure that the image was of an appropriate size for the latent space to handle [16].

The encoded transformation is then passed through a custom transformation layer that applies a CDF 97 discrete wavelet transform typically used for preserving image fidelity when compressing JPEG images [16]. In the case of penumbral images, a linear transformation is also applied. Mathematically, the CDF97 transform takes the following general form, where $f$ represents the encoded latent representation of the image:

$$\tilde{f} = \mathrm{IDWT}^{\mathrm{L}}\left(\mathrm{DWT}^{\mathrm{L}}(f)\right) \approx f \tag{3}$$

L stands for the number of times the transform and its inverse are applied to the image in the latent space. The optimal number of levels was found to be 2, with further levels yielding diminishing returns for results. IDWT refers to the inverse wavelet transform, which is usually applied after performing a discrete wavelet transform (DWT) on an array.

All the synthesis and analysis filter coefficient arrays were initialized as 9-tap filters. The analysis high pass and synthesis lowpass filters were padded by zero at the $0^{th}$ and $8^{th}$ positions to replicate appropriate CDF97 behavior as described in the original Daubechies publication. The analysis low-pass and synthesis high-pass filters, on the other hand, are not padded and contain all 9 taps like so:

$$\begin{aligned} h[n] &= \{h_0, h_1, h_2, h_3, h_4, h_5, h_6, h_7, h_8\} \\ h[n]_{\mathrm{padded}} &= \{0, h_1, h_2, h_3, h_4, h_5, h_6, h_7, 0\} \end{aligned} \tag{4}$$

This aligns with the boundary handling method described in the original implementation of the CDF 9/7 biorthogonal wavelet filter [14,18]. The numerical values of the nonzero coefficients were derived using the lifting scheme and scaling normalization steps developed by Daubechies [14]. When a linear transformation is needed, the image goes through the following changes in the latent space before being passed to the CDF97 wavelet transform, where b is the batch index, c is the channel index, H is the height index, and W is the width index:

$$\begin{aligned} y_{b,c,1,1} &= \frac{1}{H \times W} \sum_{i=1}^{H} \sum_{j=1}^{W} x_{b,c,i,j} \\ x_{b,c,1,1} &\rightarrow x_{b,c} \\ z_b &= W x_b + b \\ {y'}_b &= W' z_b + b' \end{aligned} \tag{5}$$

These transformations stabilize the variance and reduce the impact of noise in image data. Moreover, the latent space keeps the image vectors (32, 32) in dimension. The feature maps stay constant as well.

After the data samples are passed through the latent space, the decoder reconstructs the original image from the compressed representation generated from the custom layer. It consists of seven convolutional layers. The first two are followed by a batch normalization and leaky ReLU layer. The third and fourth layers are only superseded by leaky ReLU activation functions. To prevent checkerboarding in the image background during reconstruction, the last three layers are preceded by up-sampling layers in bicubic mode. Each time, the image is up-sampled by a factor of 2 using bicubic interpolation, which doubles the spatial dimensions at each step and reverses the down-sampling done by the encoder. While the first five activation layers are ReLU functions, the activation function of the last layer is a Sigmoid outputting values between 0 and 1. This makes it optimal for reconstructing and working with grayscale images that typically have pixel values between 0 and 1.

The images were run through the autoencoder for 30 epochs. The learning rate was 0.001 and data indices were shuffled each epoch. A batch size of 10% of the sample size was found to be the most optimal for reconstruction. Hence, for the simulated ground truth data used for benchmarking, the batch size was set to be 300 and for the NIF data, the batch size was set to be 12. A smooth variant of the L1 loss, known as Smooth L1 or Huber loss, was used as the loss function. It behaves quadratically for small errors and linearly for large errors, reducing sensitivity to outliers while maintaining differentiability [19]. Smooth L1 Loss can be expressed mathematically like so:

$$L(x) = \begin{cases} 0.5(x-y)^2 & if\ |x-y| \le 0 \\ |x-y| & if\ |x-y| > 0 \end{cases} \tag{6}$$

It was weighted with the PyTorch MSSSIM function in a 7:3 ratio, respectively. This algorithm was used in lieu of commonly used alternatives such as Mean Squared Error due to its resilience to outliers and frequent use in image denoising and dehazing algorithms [20]. Additionally, when reconstruction error was compared, the model performed better with Smooth L1 Loss as opposed to MSE for the given datasets. Fig. 5 displays visual loss results across two training set sizes for the weighted method compared to other common loss functions.

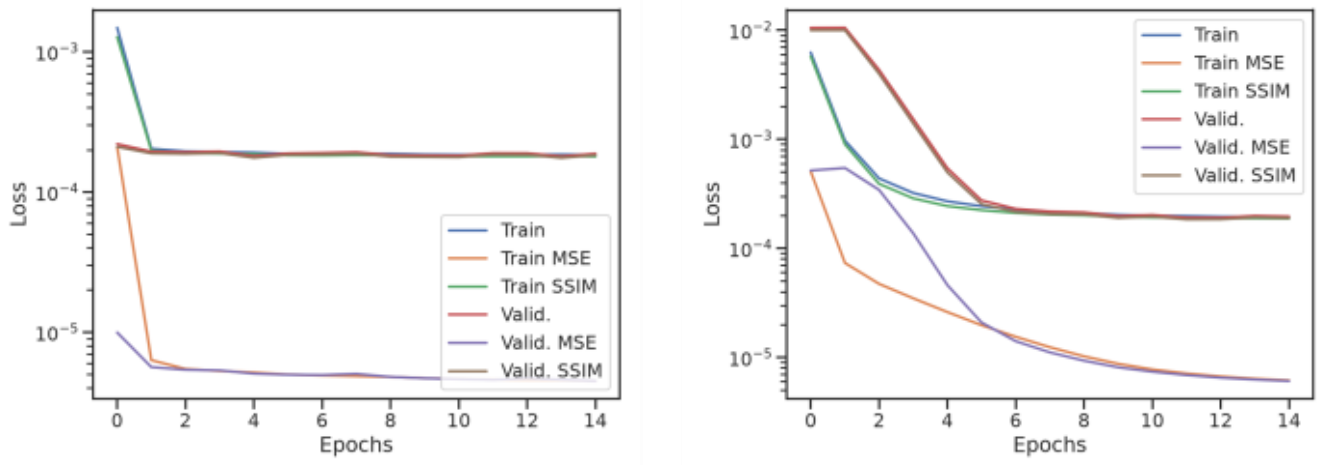

**Fig. 5.** The left-hand side of the figure illustrates loss for the model trained on 2400 images along with 600 validation holdouts. It converges soon after the first epoch because of the large volume of samples in both sets. The right-hand side depicts the model's loss behavior when trained on a smaller number of images, such as a training set of 240 images and validation set of 60 images, the loss is somewhat higher and decreases more slowly. MSE and SSIM-only losses are shown side-by-side with the combined SSIM and Smooth L1 loss curve. When compared to MSE, the combined and SSIM curves are more stable, with less significant jumps in loss and no downward concavity which often indicates exploding gradients and unstable learning rates [21]. The Smooth L1 component provides robustness to outliers while the SSIM component preserves perceptual image quality, resulting in a combined loss curve that balances both objectives.

## III. Results

Both the synthetic ground truth and NIF datasets were utilized to generate and analyze results. These datasets objectively assessed the autoencoder's ability to denoise mixed Gaussian-Poisson noise under the physics constraints imposed upon the system by the forward model. The NIF dataset was used as proof of concept to ascertain that an autoencoder model could perform generalizable denoising in practical, noisy conditions that reflect true experimental variability and Gaussian-Poisson characteristics within NIF datasets. The denoised, reconstructed results from the autoencoder were compared to datasets denoised with BM3D, a widely regarded benchmark method in image denoising [22]. For each method, the denoised reconstructions were compared to the ground truth. Assessment metrics included visual polar and cartesian coordinate visualization, radial position of the edge, spread of the edge at transitions, residuals, and radial profile. Tikhonov regularization, also known as ridge regression, was also applied along with Weiner deconvolution to evaluate how effectively the reconstruction recovered the original source when compared to the ground truth.

To create a reliable denoising solution for the images at hand, it was essential to validate that the noise present in the NIF data was indeed mixed Gaussian-Poisson in nature [7]. As an alternative, histogram analysis, local variance analysis, and power spectrum analysis as displayed in Fig. 6 were employed to acquire more information about the nature of the noise present in the data [23].

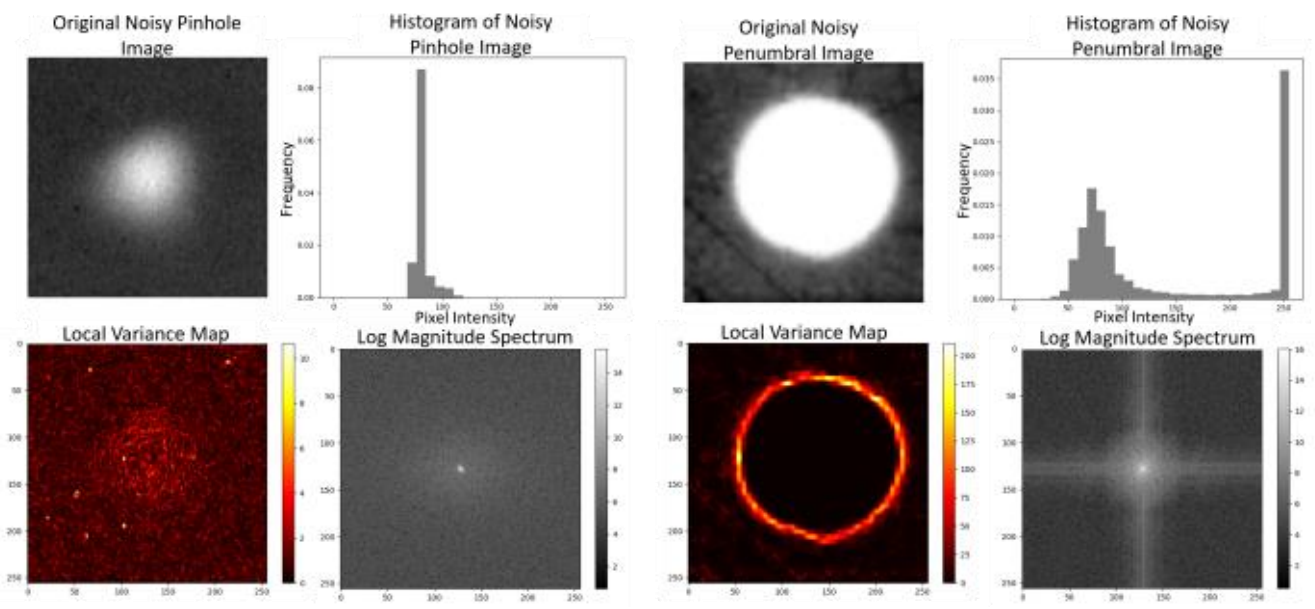

Fig. 6: To confirm the type of noise present in the images, a histogram, local variance plot, and power spectra were generated for a pinhole (left) and penumbral (right) image from the N210808 shot. The noise was ascertained to be mixed Gaussian and Poisson across the images due to the features and characteristics of the plots.

The results of these analyses for both pinhole and penumbral images indicate mixed Gaussian-Poisson noise. Pixel intensity histograms of NIF data are slightly skewed left with a bell-curve shape in both cases. This shape suggests the presence of Gaussian noise, while the skew indicates that Poisson noise may also contribute to visual distortions in the image [23]. In the case of the pinhole image, the local variance map displays higher variance in lighter areas of the image when compared to darker areas. A similar trend is seen at the edges of the penumbral image. Poisson noise in low-light conditions is often characterized as greater in magnitude at higher pixel intensities; hence, increased variance at higher pixel intensities in both images is an indicator of this type of noise [24]. In the case of the pinhole image, Fourier spectrum analysis results in a grainy background with poorly defined frequency and phase components. The background indicates that the image contains high levels of Gaussian noise, while the lack of frequency and phase bands in the case of the pinhole aperture indicate the presence of Poisson noise as well due to its signal dependent nature [25,26]. The penumbra, on the other hand, displays strong frequency and phase components against a grainy background. This indicates that while Poisson noise may not be as significant a problem in penumbral images, Gaussian noise is still pervasive throughout this portion of the image.

Reconstruction error measures the discrepancy between an autoencoder's input and its reconstructed output, often quantified using pixel-wise losses and structural similarity metrics. During training, the model's weights and biases are iteratively updated to minimize this error [27]. A viable autoencoder, in reconstructing noisy data, must be able to produce a result that matches the original ground truth image the inputs are derived from as closely as possible. While the most common method for assessing reconstruction fidelity is Mean Squared Error (MSE), other values may be included such as Peak Signal to Noise Ratio, Structural Similarity Index, and edge preservation metrics [27,28]. Since the most significant aspect of reconstruction for neutron imaging is edge preservation and source reconstruction, most of the results are focused on this. Visual polar and cartesian coordinate visualization, radial position of the edge, spread of the edge at transitions, residuals, and radial profile were all considered. Additionally, Tikhonov regularization was also applied to evaluate how effectively the reconstruction recovered the original source.

Fig. 7 displays a polar and cartesian visualization of the reconstructions indicated high fidelity between the ground truth and the denoised images. Key structural features and spatial patterns were preserved across both representations, indicating that the denoising process maintained the essential image characteristics without introducing noticeable artifacts. The alignment between the ground truth and denoised images was particularly evident in the preservation of the Gaussian blur along the edges. This suggests that the autoencoder effectively distinguishes between the blur inherent to the ground truth and extraneous noise introduced during noise generation in both the Gaussian and Gaussian-Poisson cases. This selective retention further supports the qualitative accuracy of the reconstruction.

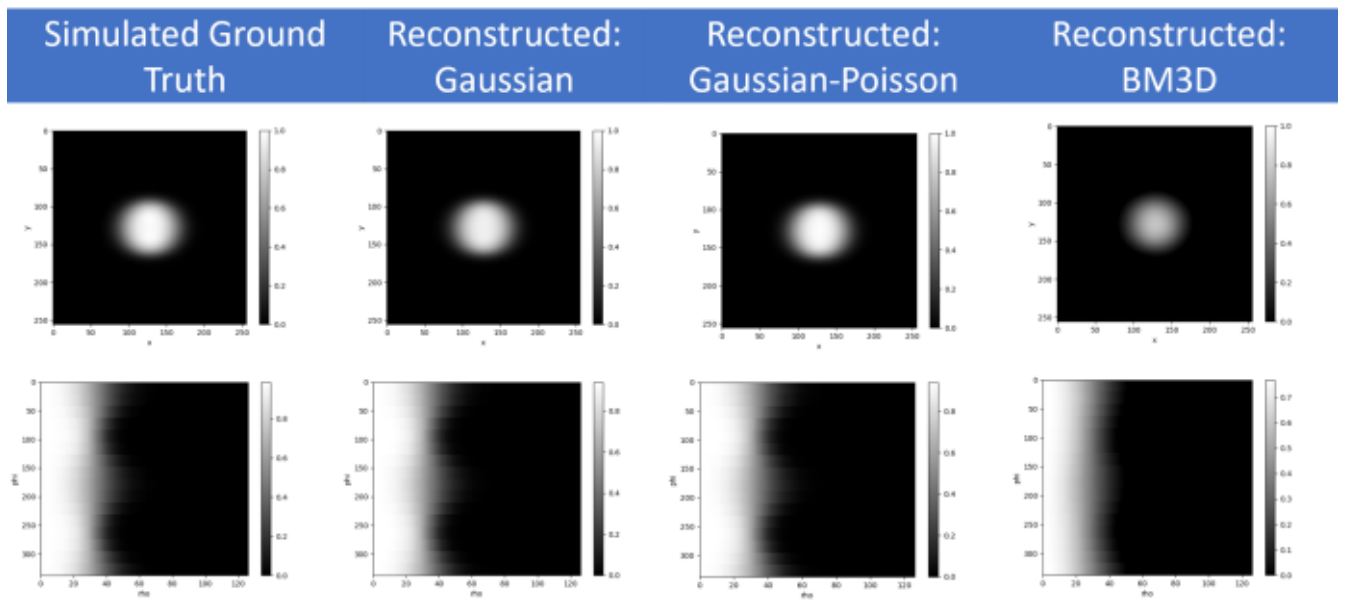

Fig. 7: The simulated ground truth visualized side-by-side with autoencoder reconstructions of images corrupted by Gaussian noise and Gaussian-Poisson noise. For comparison, the BM3D method was utilized to reconstruct an image with Gaussian-Poisson noise and can be seen on the right. Both cartesian (top) and polar (bottom) results indicate higher fidelity between the ground truth and autoencoder reconstructions than compared to the ground truth and BM3D reconstruction.

Fig. 8a illustrates a quantitative analysis of the radius as a function of degree in the simulated ground truth, compared to autoencoder-based reconstructions under Gaussian and Gaussian-Poisson noise models, demonstrated performance on par with established denoising techniques such as BM3D. In images corrupted by Gaussian noise, the autoencoder reconstructions exhibited near-perfect agreement with the ground truth, accurately recovering the underlying magnitude of the radial profile. In the presence of Gaussian-Poisson noise, the reconstructions displayed a marginal reduction in estimated radius of approximately four pixels but effectively preserved the overall radial trend consistent with the simulated ground truth.

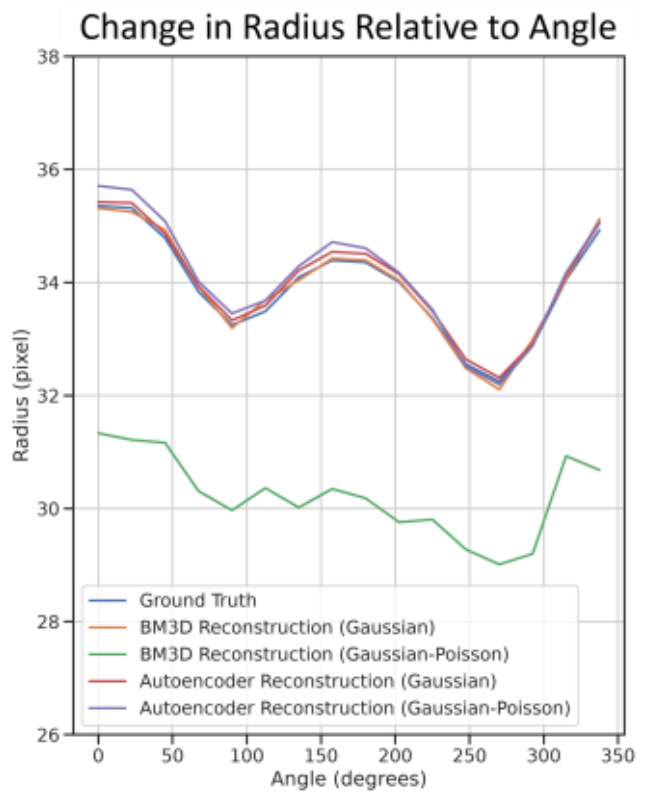

Fig. 8a: Change in radius relative to angle for reconstructed images. Ground truth is compared to autoencoder and BM3D reconstructions graphically.

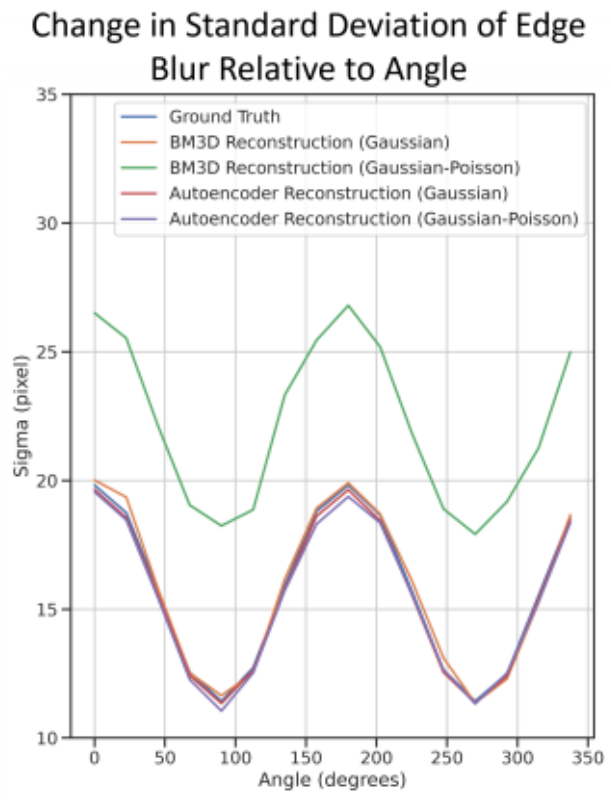

Fig. 8b: Change in standard deviation of image edge blur in pixels relative to angle. Ground truth is compared to autoencoder and BM3D reconstructions graphically.

To further compare edge fidelity across differing denoising methods and noise models, the absolute radius error and relative fidelity scores were computed like so:

$$\varepsilon_r = |\hat{r} - r_{GT}| \tag{7}$$

$$Fidelity_{AE\ vs.BM3D} = \left(1 - \frac{{\varepsilon_r}^{AE}}{{\varepsilon_r}^{BM3D}}\right) \times 100 \tag{8}$$

For samples corrupted by Gaussian noise, the autoencoder reconstructions showed a mean absolute radius error of 1.447 pixels, outperforming BM3D's 1.632 pixels. The relative fidelity scores were computed to be 95.93% and 95.41%, respectively. For samples corrupted by Gaussian-Poisson noise, autoencoder reconstructions showed a mean absolute radius error of 4.670 pixels, while BM3D reconstructions had an error of 5.001 pixels. For these sample sets, the relative fidelity scores were 86.87% for autoencoder reconstructions and 85.94% for BM3D reconstructions. These results demonstrate that the autoencoder preserves radial structure with high accuracy under both noise models and performs competitively with traditional denoising methods, particularly in terms of structural and metric fidelity.

Fig. 8b shows a visualization of the Gaussian blur variance, also known as $\sigma$, which at edge transitions in the reconstructions exhibited high fidelity to the ground truth in images degraded by Gaussian noise, accurately preserving edge sharpness and gradient continuity. This is supported by a high edge PSNR of 34.2 dB and a low gradient magnitude similarity deviation (GMSD) of 0.078. Under Gaussian-Poisson noise, the reconstructions exhibited elevated variance levels, reflecting the autoencoder's increased difficulty in disentangling structured blur from unnecessary noise under more complex noise conditions. This is supported by an elevated local gradient variance at the edges, rising from 9.0406 to 9.9690. Despite this challenge, the reconstructions maintained the expected σ-blur trend, albeit at a higher variance overall. Autoencoder-based methods outperformed BM3D in preserving the integrity of the σ profile in both Gaussian and Gaussian-Poisson noise situations, demonstrating superior retention of spatially dependent blur characteristics when compared to conventional filtering methods. Specifically, the mean squared error (MSE) between the reconstructed and ground truth σ-blur profiles was 0.015108 for the autoencoder, compared to 0.032195 for BM3D, indicating significantly improved spatial fidelity in blur retention. Additionally, edge PSNR and GMSD scores consistently favored the autoencoder, underscoring its superior ability to preserve spatially dependent blur characteristics compared to conventional filtering approaches.

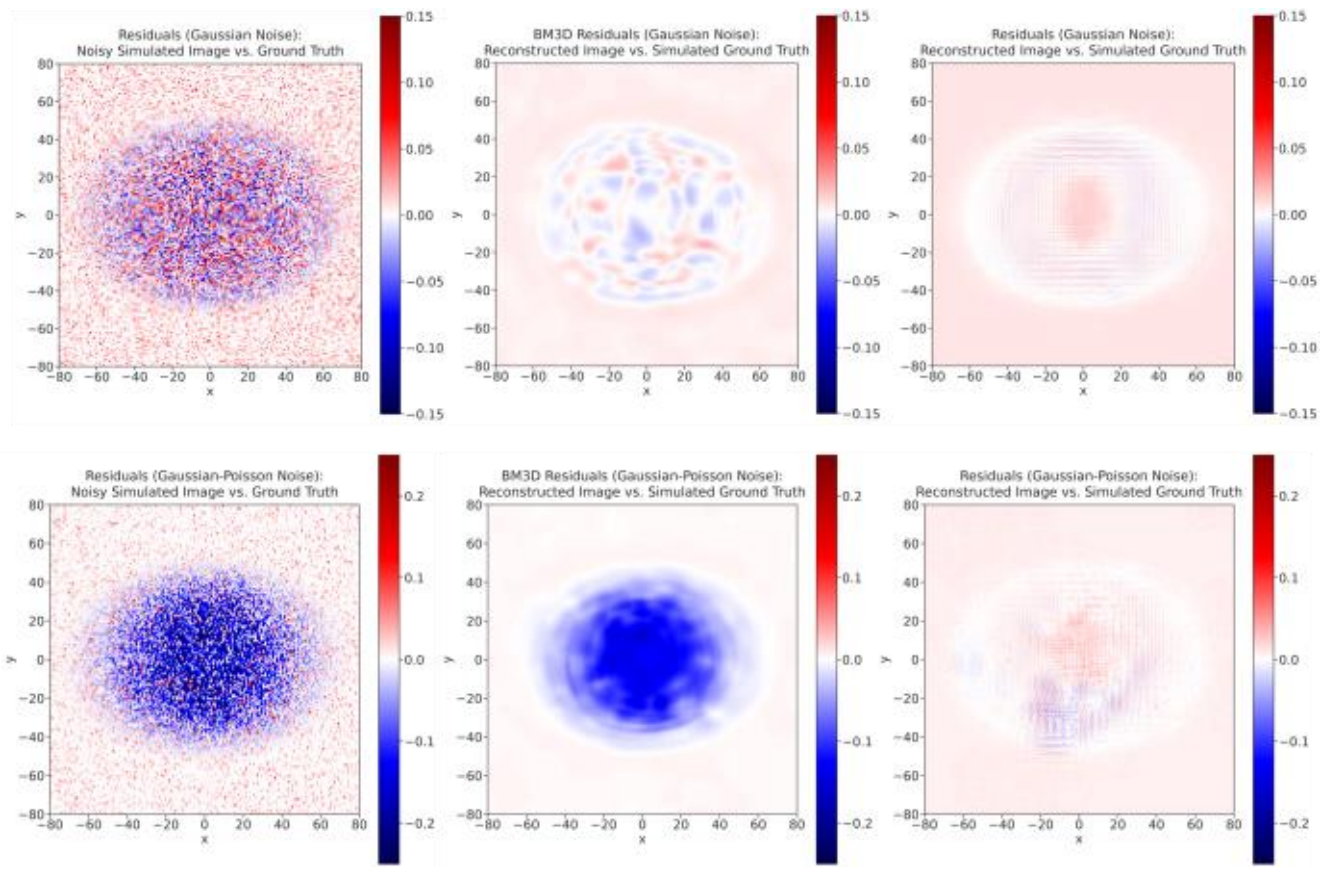

Fig. 9: Residual analysis of reconstructions relative to the ground truth indicates high fidelity in the presence of Gaussian noise. Autoencoder reconstructions outperformed BM3D reconstructions in the presence of mixed Gaussian-Poisson noise.

Fig. 9 displays residual analysis of the reconstructions relative to the ground truth. It demonstrated high fidelity in the presence of Gaussian noise. Residuals in this case were tightly distributed, with a mean of 0.012, a median of 0.011, and a standard deviation of 0.005. Residual values were consistently constrained within the range [0.00, 0.03] across the image domain. For images affected by Gaussian-Poisson noise, residuals were notably higher and more dispersed, with a mean of 0.041, a median of 0.038, and a standard deviation of 0.018. They generally ranged between [0.00, 0.10], reflecting the increased complexity in accurately reconstructing pixel intensities under mixed noise conditions. However, the autoencoder was able to reduce residuals of the reconstructions by 65.8% on average in the case of images corrupted by Gaussian-Poisson noise and by 69.3% on average in the case of images corrupted by Gaussian noise alone. This is an improvement when compared to BM3D, which only reduces residuals by 46.3% on average in the case of images with Gaussian-Poisson noise and 50.8% in the case of images with Gaussian noise alone. In the case of BM3D, the average residual values of 0.059 (mean), 0.055 (median), and 0.022 (standard deviation) under Gaussian-Poisson noise, and 0.021 (mean), 0.020 (median), and 0.007 (standard deviation) for Gaussian noise demonstrate the autoencoder's improved denoising capability when compared to BM3D in the presence of complex, nonlinear noise.

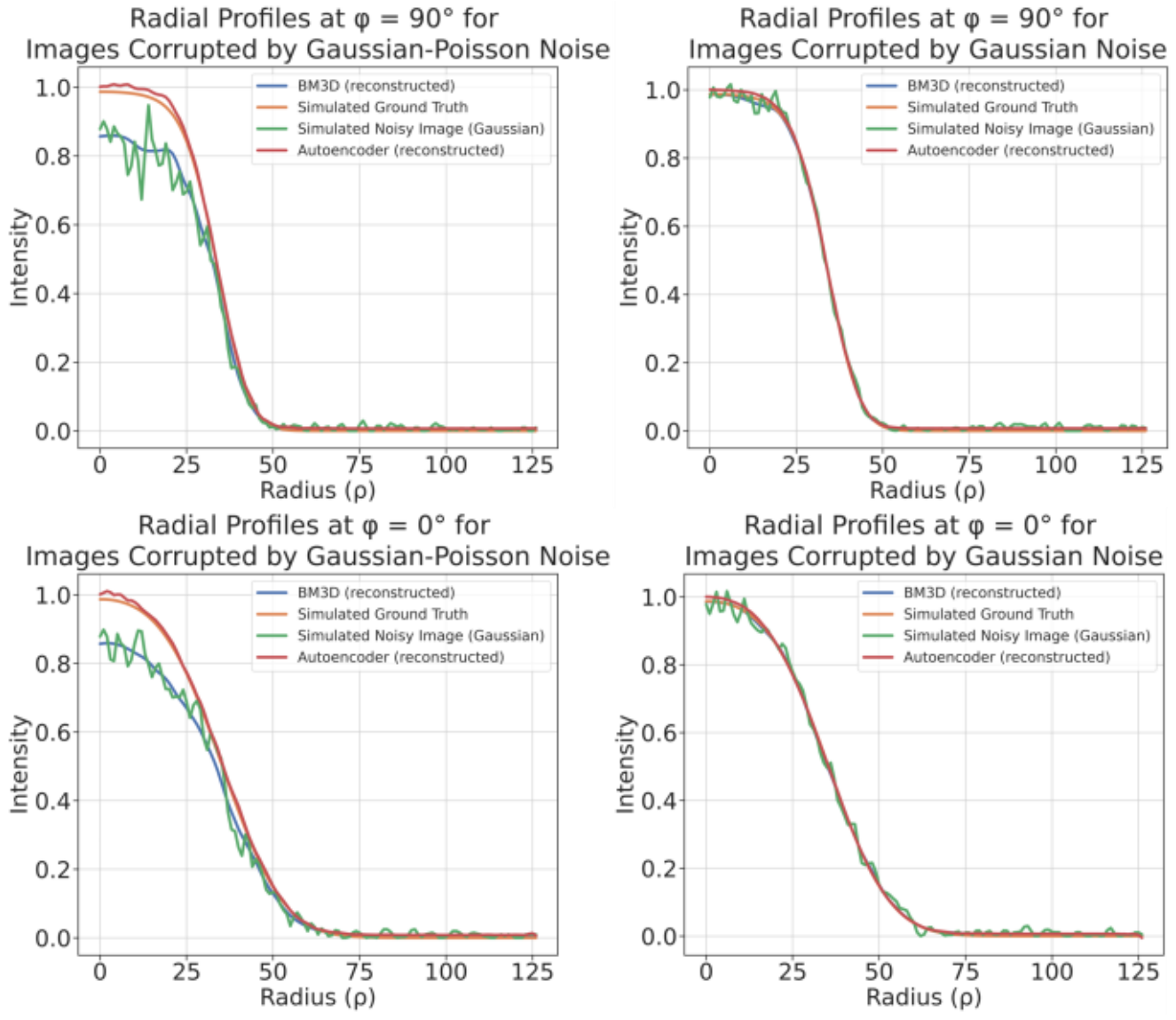

Fig. 10: Radial pixel intensity profiles of the circular pinhole image extracted along 0° and 90° directions. The ground truth, noisy input, autoencoder reconstruction, and BM3D denoising images are compared on the same plot for each angle. Profiles demonstrate the denoising performance and edge preservation of each method.

As shown in Fig. 10, radial profile analysis of the autoencoder reconstructions relative to the simulated ground truth revealed that, under Gaussian-Poisson noise, the reconstructed pixel intensities were initially attenuated, with values starting around 0.8 compared to the ground truth peak of 1.0. However, as the radius increased, the profile gradually converged toward the ground truth, indicating that the reconstruction effectively recovered the overall structure despite initial underestimation. In contrast, reconstructions from Gaussian noise exhibited near-perfect alignment with the ground truth across the entire radial domain, demonstrating the autoencoder's high fidelity under simpler noise conditions.

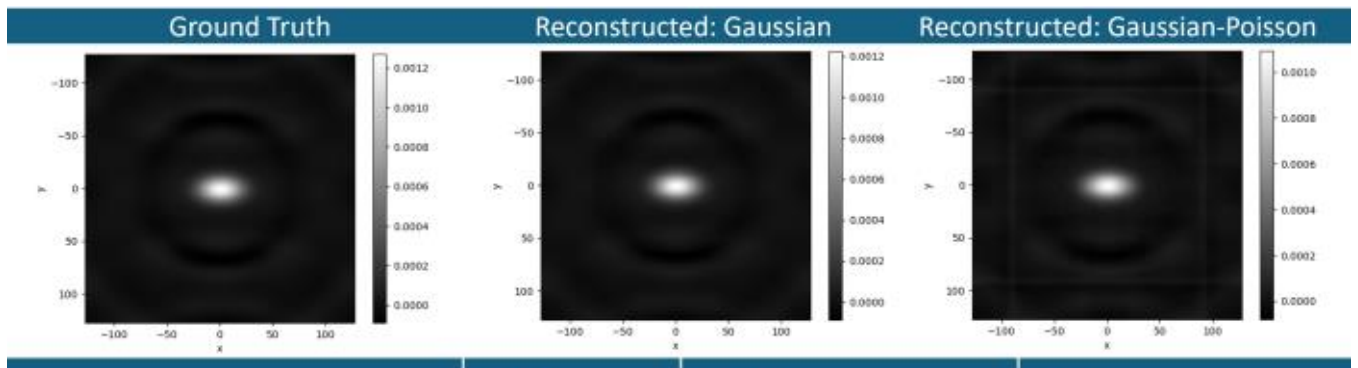

| | SSIM | PSNR | MSE |
|---|---|---|---|
| Gaussian | $0.99 \pm 2.22^{-12}$ | $99.88 \pm 1.02^{-5}$ | $1.03^{-10} \pm 2.41^{-16}$ |
| Gaussian-Poisson | $0.99 \pm 2.66^{-7}$ | $95.79 \pm 1.28^{-5}$ | $3.45^{-10} \pm 7.29^{-12}$ |

Fig. 11: The recovered source profile obtained by applying regularized Wiener filtering to deconvolve the disk point spread function from the observed image. The reconstruction successfully recovers the underlying 2D Gaussian source with regularization parameter α = 0.01 and estimated SNR = 10.

To validate the fidelity of the reconstructions further, the source image was recovered from both the ground truth and reconstructed outputs using Wiener deconvolution in conjunction with Tikhonov regularization, incorporating the known point spread function (PSF) of the forward model defined as the sum of the initial disc before convolution with the source of the forward model. In both the Gaussian and Gaussian-Poisson noise scenarios, the deconvolved reconstructions closely approximated the true source, demonstrating the autoencoder's ability to preserve critical spatial information through the noise. The recovered sources exhibited high structural similarity to the ground truth, further validating the effectiveness of the reconstruction pipeline under varying noise conditions. Fig. 11 shows a visualization of the results relative to the ground truth along with structural similarity, peak signal to noise ratios, and mean squared error values. After training and validating the autoencoder using simulated ground truth data, the model was applied to NIF pinhole and penumbral images. Fig. 12 displays a post-training result of running the autoencoder on 50 pinhole images and 6 penumbral images with a batch size of 6 and 2, respectively. The reconstructions exhibited strong qualitative and quantitative agreement with expected structural features, indicating that the autoencoder successfully generalized from simulated to real-world conditions. This demonstrates the model's robustness and practical applicability in processing experimental data following tuning on synthetic datasets.

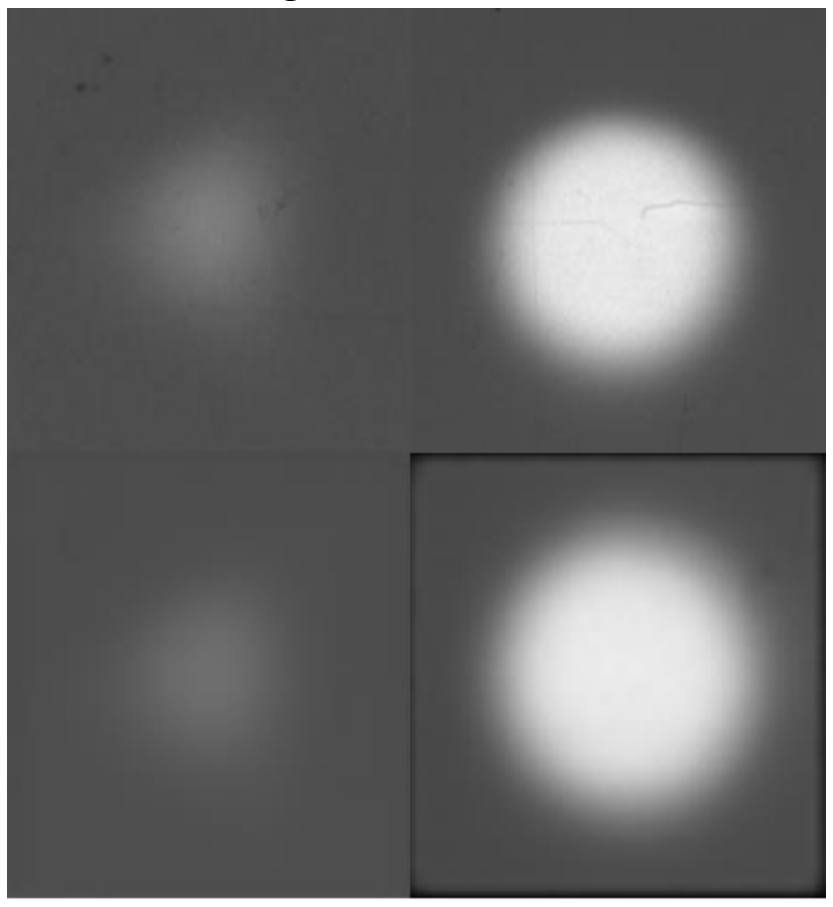

Fig. 12: Denoised NIF pinhole (left column) and penumbral images (right column), respectively. These results were achieved with the benchmarked autoencoder on a limited amount of training data. They indicate that the model has consistent reconstruction capability in real-world scenarios where thousands of data points may not be accessible.

## IV. Discussion

Our study found that an unsupervised machine learning approach outperforms conventional methods when reconstructing images corrupted with Gaussian and Gaussian-Poisson noise in physics models. The observed improvements in reconstruction quality and edge preservation suggest that unsupervised learning frameworks offer a viable and scalable alternative in the denoising work necessary to optimize operations in computational physics. These methods can autonomously learn from intrinsic data features, reducing reliance on external training datasets and potentially simplifying implementation across diverse noise models. Results indicate that an unsupervised autoencoder model is particularly effective in preserving image edges when reconstructing samples corrupted by Gaussian noise.

For images affected by Gaussian-Poisson noise, the model achieved performance comparable to conventional methods. Most prior studies in image reconstruction of samples corrupted by Gaussian and Gaussian-Poisson noise in ICF imaging have relied on supervised learning frameworks which require large, labeled training datasets [6]**.** In contrast, this work's unsupervised autoencoder approach operates without external labels and learns directly from the noisy input data. While supervised models have achieved strong results, unsupervised models have the potential for superior performance in edge preservation under Gaussian and Gaussian-Poisson noise above that of conventional methods, a particularly important metric to address in physics contexts where low-noise ground truth data is unavailable. Unsupervised frameworks can therefore provide viable, data-efficient alternatives that more closely match reality.

These results underscore the utility of unsupervised learning as a powerful tool in denoising neutron imaging data for ICF research. The autoencoder framework presented here not only eliminates the need for large, labeled datasets but also reduces the complexity of next-generation preprocessing pipelines, such as multi-step transformations often used for Gaussian-Poisson denoising. Additionally, it provides an initial step in the process of using AI for image reconstruction to remove the aperture function in neutron imaging. This reduction in complexity is particularly valuable in low-throughput, data-sparse environments like the NIF, where time-resolved images are scarce and noisy. Unsupervised image processing is advantageous for adjusting experimental parameters on the fly without an over-reliance on simulated data and parameters. Additionally, by incorporating a CDF 97 wavelet transform in the latent space, the network gains sensitivity to fine-grained features across multiple scales—an essential quality for detecting subtle spatial patterns and edge information in neutron source distributions. Wavelet-based representations appear to complement and enhance the autoencoder's ability to generalize, offering a more robust encoding of structural features that conventional filtering methods tend to degrade.

## V. Conclusion

This study demonstrates the effectiveness of using an unsupervised autoencoder constructed with custom and convolutional layers for the denoising of NIF ICF pinhole images. This method outperforms widely utilized filtering methods such as BM3D, Wiener filtering, and Gaussian filtering when minimizing reconstruction error and maximizing edge preservation. Additionally, it decreases the reliance of the model on simulated data past validation of its efficacy when compared to supervised machine learning denoising methods that always require a training set that often needs to be simulated due to a lack of ground truth data. The custom layer in the bottleneck of the autoencoder contributes to better denoising performance when compared to baseline models without it. Improving unsupervised denoising methods is essential for better visualizing ICF events, as a lot of images are taken in low-light conditions where it is difficult to find a ground truth or even have a large training set of real data. Autoencoders perform denoising by relying solely on features extracted directly from the input image, rather than external training datasets. Often, they rely solely on the image itself during denoising, with the denoising process being based on key features extracted from the image currently being processed by the autoencoder. It is directly informed by the key features of the image at hand as opposed to arbitrary or simulated ground truth training datasets. Unsupervised methods are more adaptable to real-world scenarios like ICF, where acquiring ground truth experimentally is not feasible. However, it is imperative to quantify overfitting and ensure that the boundaries for the algorithm's utility are established. Future work would involve addressing the limitations of the current methodology. Currently, the model has primarily been tested on neutron images and has not been validated to be generalizable to other types of images such as x-ray images. It has also not been rigorously validated with various source shapes and apertures that are utilized less often. Additionally, the model is not generalized to large pixel spaces and the images being fed into it are currently just smaller segments of a wider neutron imaging plate. Finally, despite its strengths, limitations remain in the unsupervised model's ability to fully resolve nonlinearities introduced by compound noise distributions such as overlapping Poisson-Gaussian processes. Future work should explore the integration of physics-informed loss functions or hybrid models that combine unsupervised and semi-supervised paradigms to further improve noise-specific adaptability. Exploring domain-specific augmentations or incorporating simulation-based priors may also enhance model performance, particularly in low-signal or anisotropic imaging scenarios typical of early-stage ICF implosions. Ultimately, the findings from this study reinforce the potential of unsupervised machine learning for neutron image enhancement and open the door to a broader class of self-tuning algorithms for ICF diagnostics. By bridging the gap between traditional filtering approaches and data-intensive supervised learning, this methodology presents a proof-of-concept for a scalable, generalizable framework for future fusion research applications, including time-series denoising, 3D tomographic reconstruction, and anomaly detection in neutron yield patterns. In a field where ground truth is not easily available, the applicability of unsupervised methods in this space is explored through this study.